\documentclass[11pt,twocolumn]{article}

\usepackage[T1]{fontenc}
\usepackage[utf8]{inputenc}
\usepackage{times}
\usepackage{microtype}
\usepackage{amsmath}
\usepackage[top=2.5cm,bottom=2.5cm,left=1.8cm,right=1.8cm,columnsep=0.6cm]{geometry}
\usepackage{booktabs}
\usepackage{multirow}
\usepackage{array}
\usepackage[colorlinks=true,linkcolor=black,citecolor=black,urlcolor=blue]{hyperref}
\usepackage{url}
\usepackage{natbib}
\usepackage{titlesec}
\usepackage{caption}
\usepackage[bottom]{footmisc}
\usepackage{orcidlink}
\usepackage{appendix}

\setcitestyle{authoryear,open={(},close={)}}

\titleformat{\section}{\large\bfseries}{\thesection}{0.5em}{}
\titleformat{\subsection}{\normalsize\bfseries}{\thesubsection}{0.5em}{}
\titlespacing*{\section}{0pt}{8pt}{3pt}
\titlespacing*{\subsection}{0pt}{6pt}{2pt}
\newcommand{\code}[1]{\texttt{\small #1}}

\begin{document}

\twocolumn[{%
\begin{center}
{\LARGE\bfseries The Sleeping Agent: What Gist-Based Context\\[4pt]
Compression Loses and Why}\\[10pt]
{\normalsize N. E. Kyrkewood\,\orcidlink{0009-0004-4763-6134} (Independent Researcher)}\\[4pt]
{\small \url{https://github.com/kyrkewood/sleeping-agent}}
\end{center}
\vspace{4pt}
\begin{quote}
\small\textbf{Abstract.}\enspace
Gist-based context compression---summarising older conversation history into compact
representations---is a common approach in long-horizon language model agents, yet its
effect on different types of memory retrieval is poorly understood.
We use Salience-Weighted Consolidation (SWC), a biologically-inspired compression
framework motivated by sleep-based memory consolidation, as a diagnostic probe to study
when gist compression helps and when it hurts.
SWC scores conversation history by salience, partitions it into priority tiers, and
applies structured gist abstraction to mid-priority content.
Evaluating four conditions on all ten LoCoMo conversations---1\,935 matched text-only
questions in total, 1\,501 used in the primary aggregate after excluding Category~5
(adversarial) questions---at temperature~0, we find a consistent task-type interaction:
gist compression substantially outperforms truncation on multi-hop reasoning and
single-hop factual questions, but temporal questions remain substantially harder under
compression, with compressed conditions scoring well below the full-context reference
on the conversations where both are evaluated.
We trace this failure to a specific mechanism: the gist abstraction prompt preserves
relational and event structure while discarding dates and times.
A preservation analysis across all ten conversations confirms the mechanism: an
approximately 20-fold increase in temporal expression preservation
(3.05\%$\to$62.39\%) with a one-sentence prompt modification, while named entity and
event preservation rates barely change ($\times$1.02 and $\times$1.11), demonstrating
that the fix is a precision instrument.
The prompt modification recovers $+0.314$ $[0.254, 0.375]$ judge accuracy on
category-2 (temporal) questions in the matched set.
Code and results: \url{https://github.com/kyrkewood/sleeping-agent}.
\end{quote}
\vspace{8pt}
}]

\section{Introduction}

Memory management in long-horizon language model agents involves a recurring design
decision: when conversation history grows too long to fit in the active context window,
what should be kept, compressed, or discarded?
A common practical approach is reactive summarisation: when context exceeds a
threshold, older history is condensed by a summarisation model and substituted back.
This applies compression uniformly, regardless of what the history contains or what
questions the agent will later need to answer.

Recent work has shown that structured memory systems substantially outperform simple
compression. LightMem \citep{fang2025lightmem}, Mem0 \citep{chhikara2025mem0},
MemMachine \citep{wang2026memmachine}, and related systems report substantially higher
LoCoMo scores than truncation and sliding window, often in the 0.7--0.9 range, though
results come from systems of varying architectural complexity and evidence quality
ranging from peer-reviewed (LightMem, ICLR~2026) to self-reported preprints.
These systems represent a clear advance, but their complexity makes it difficult to
understand which components of memory management matter and why.

We use Salience-Weighted Consolidation (SWC)---a biologically-inspired compression
framework---as a diagnostic probe.
SWC scores context chunks by estimated salience, partitions them into priority tiers,
and applies structured gist abstraction to mid-priority content.
We identify a consistent task-type interaction: gist compression outperforms truncation
on multi-hop and single-hop questions but fails specifically on temporal ones.
A preservation analysis confirms the mechanism: a temporal-protecting prompt
modification causes an approximately 20-fold increase in temporal preservation while
barely affecting named entity and event preservation ($\times$1.02 and $\times$1.11).
That single prompt change recovers $+0.314$ judge accuracy on category-2 (temporal)
questions in the matched set, with the direction positive in all ten conversations.
The failure mode is a correctable design omission.

The finding that temporal anchors are systematically and selectively lost under generic
gist compression suggests a design principle for gist-compression components across
memory architectures: temporal anchors should be explicitly protected rather than left
to generic abstraction prompts.

\section{Background}

\subsection{Two-Stage Model and Selective Replay}

The two-stage model \citep{mcclelland1995complementary} proposes that memories are
initially encoded in the hippocampus and gradually transferred to the neocortex during
sleep via selective replay. Memories associated with reward, novelty, or high salience
are preferentially reactivated.

\subsection{Forgetting as Function}

The synaptic homeostasis hypothesis \citep{tononi2006sleep} proposes that sleep reverses
waking-induced synaptic potentiation through global downscaling, preserving stronger
synapses while pruning weaker ones. Forgetting is an active noise-reduction process.

\subsection{Episodic Specificity and Gist}

Sleep converts episodic memories (specific events, times, contexts) to more semantic
representations. This conversion is adaptive for long-term knowledge but lossy for
specific detail. Named entities and event mentions survive at higher verbatim rates than
temporal expressions in generic gist summaries ($\sim$8\% and $\sim$5\% versus
$\sim$3\%). The operative distinction is \emph{temporal versus non-temporal}, not
episodic versus semantic.

\section{Related Work}

\subsection{Context Management}

The `lost in the middle' phenomenon documents performance degradation as context length
increases \citep{liu2024lost}. Proactive interference is an additional source of
degradation \citep{wang2025unable}. Truncation is recency-biased; sliding window
summarisation applies uniform compression regardless of content importance.

\subsection{Memory Systems}

LightMem \citep{fang2025lightmem} reports substantial accuracy and token-efficiency
gains on LoCoMo using a staged architecture with sleep-time offline update.
Mem0 \citep{chhikara2025mem0} uses dynamic extraction, consolidation, and retrieval
with a graph-memory variant, reporting a 26\% relative LLM-as-judge improvement over
OpenAI on LoCoMo. MemMachine \citep{wang2026memmachine} reaches 0.917 on LoCoMo using
ground-truth-preserving episodic storage with contextualised retrieval.

SWC is not proposed as a competitor to these systems but as a controlled probe to study
gist compression effects in isolation.

SleepGate \citep{xie2026sleepgate} uses learned KV cache gating to target proactive
interference architecturally. Adaptive Focus Memory \citep{cruz2025afm} uses
multi-fidelity context packing based on semantic relevance and temporal decay.
ACON \citep{kang2025acon} optimises context compression through natural language
feedback. ENGRAM \citep{patel2025engram} proposes lightweight memory orchestration
using structured extraction and retrieval.

\section{Salience-Weighted Consolidation}

\subsection{Design Principles}

SWC draws three principles from sleep neuroscience: \textbf{proactive scheduling}
(consolidation at idle windows, not capacity limits), \textbf{salience-weighted
retention} (chunks scored and partitioned), and \textbf{staged compression}
(salience-scoring pass followed by gist abstraction).

\subsection{Stage 1: Salience Scoring}

Each session chunk receives a score from three signals:

\textbf{Downstream similarity (0.4):} Cosine similarity between chunk and subsequent
turn embeddings, computed locally using \code{all-MiniLM-L6-v2} via
\code{sentence-transformers} (no embedding API called; only Anthropic APIs used for
generation and judging). Sessions with temporally specific facts may score low despite
later relevance---consistent with the temporal failure mode.

\textbf{Recency (0.3):} Session index $\div$ total sessions.

\textbf{Information density (0.3):} Length-normalised named entity and number count.

Scores partition chunks: high priority ($\geq$0.6, verbatim), mid priority
(0.3--0.6, gist-compressed), low priority ($<$0.3, discarded). The two most recent
sessions and session~0 are always high priority. If history exceeds the 4\,000-token
budget, high-priority chunks are dropped lowest-salience-first.

\subsection{Stage 2: Gist Abstraction --- SWC-Full}\label{sec:swcfull}

Mid-priority chunks are compressed using Claude Haiku
(\code{claude-haiku-4-5-20251001}, temperature~0):
\begin{quote}\small\itshape
Preserve: factual information about the speakers, causal relationships, decisions and
their rationale, open commitments or plans. Discard: small talk, repeated information,
emotional reactions without factual content. Output a concise paragraph of 3--5
sentences maximum.
\end{quote}
No instruction retains dates or times; the preservation analysis confirms that only
3.05\% of temporal expressions survive verbatim.

\subsection{Stage 2: Gist Abstraction --- SWC-Temporal}\label{sec:swct}

Identical to SWC-Full except for one added sentence at the prompt start:
\begin{quote}\small\itshape
You MUST preserve verbatim: all specific dates, times, durations, ages, and temporal
expressions (e.g.\ `7 May', `last Tuesday', `at 9pm').
\end{quote}
SWC-Temporal uses a separate REM cache; salience scoring is identical to SWC-Full.

\subsection{Scheduling}

SWC-Full uses salience-pressure and idle-window triggering with consolidation every
three sessions. SWC-Reactive (token-threshold only) produced near-identical scores in
early runs, consistent with offline benchmarks providing no genuine idle windows.
Scheduling is designed for online deployment and is not evaluated here
(Appendix~\ref{app:reactive}).

\smallskip\noindent\textbf{Implementation:} Salience scoring runs on commodity CPU
hardware,\footnote{MacBook Air, Apple M-series chip.} taking 30\,min--2.5\,h per
conversation.

\section{Experimental Setup}

\subsection{Benchmark and Coverage}

LoCoMo \citep{maharana2024evaluating} is the standard benchmark for long-term
conversational memory evaluation, used by the related work we compare against.
It comprises 10 multi-session dialogues averaging $\sim$600 turns and 16\,000 tokens
across up to 32 sessions. We evaluate the text-only subset; \code{locomo10.json}
contains 1\,986 QA pairs, of which 1\,935 remain after excluding image/video questions
with a keyword filter (51 excluded).

Of these 1\,935 text-only questions, 1\,501 are categories~1--4 and form the primary
aggregate (Table~\ref{tab:aggregate}); the remaining 434 category~5 (adversarial)
questions are excluded because their binary scoring rule---awarding~1 if the model says
`no information available'---rewards content removal and conflates memory quality with
response calibration (Appendix~\ref{app:adversarial}).

The matched set---the intersection of questions answered by all four compressed
conditions---comprises 1\,935 text-only questions across all ten conversations. The
primary categories~1--4 aggregate uses 1\,501 of these. Bootstrap CIs for
Table~\ref{tab:aggregate} use the 1\,501-question subset; per-category CIs use each
category's own matched question count.

\subsection{Conditions}

\textbf{Truncation:} Most recent tokens within a 4\,000-token budget, pinning
session~0.
\textbf{Sliding window:} Condenses history exceeding the budget using Haiku instructed
to retain key facts, events, and decisions.
\textbf{SWC-Full:} Two-stage salience-scoring and gist abstraction pipeline
(\S\ref{sec:swcfull}).
\textbf{SWC-Temporal:} Same pipeline with temporal-anchor protection (\S\ref{sec:swct}).
\textbf{Full context:} Entire uncompressed dialogue (no budget constraint); performance
ceiling only, evaluated on conversations~0--1.

\subsection{Category Mapping}

The authoritative mapping comes from \code{task\_eval/evaluation.py}:
\begin{center}\small
\begin{tabular}{clp{3.0cm}}
\toprule
\textbf{ID} & \textbf{Type} & \textbf{Notes}\\
\midrule
1 & Multi-hop & Partial F1\\
2 & Temporal  & `When' questions\\
3 & Open-domain & Commonsense inference\\
4 & Single-hop & Specific facts\\
5 & Adversarial & Binary; excluded\\
\bottomrule
\end{tabular}
\end{center}

\subsection{Metrics and Models}

\textbf{LLM-judge accuracy} (primary): gold and model answers passed to Claude Haiku
at temperature~0 (prompt in Appendix~\ref{app:judge}). Validated on a 60-question
blind human sample: 95\% agreement (57/60), 100\% on category~2 (17/17), all 3
disagreements conservative (judge said NO when human said YES).

\textbf{Token F1} computed for consistency with prior work but not used in the main
analysis; values in public results files.

All results use temperature~0. A prior variance analysis at API default temperature
(50 sampled pairs, 3 reruns) found mean SD~0.047; temperature-0 reruns shifted
aggregates by $\leq$0.009.

Bootstrap CIs: 95\%, 2\,000 resamples, seed~42, from the matched 1\,935-question
dataset (\code{ci\_report\_10conv\_matched.json}); 1\,501 questions for
Table~\ref{tab:aggregate}; category-specific counts for Table~\ref{tab:bycategory};
$n=315$ for the paired category-2 delta.

QA model: Claude Sonnet~4.6 (\code{claude-sonnet-4-6}). Compression/judge: Claude
Haiku (\code{claude-haiku-4-5-20251001}). Embeddings: \code{all-MiniLM-L6-v2}.

\section{Results}

\subsection{Aggregate Results}

Table~\ref{tab:aggregate} presents results on the primary analysis set:
categories~1--4, 1\,501 questions per condition.

\begin{table}[t]
\centering\small
\caption{Aggregate judge accuracy, categories~1--4 ($N=1\,501$/condition).
Full context (convs~0--1, 299 questions): 0.706 overall, 0.745 on cat.~2;
performance ceiling only.}
\label{tab:aggregate}
\begin{tabular}{lccc}
\toprule
\textbf{Condition} & \textbf{Acc} & \textbf{95\% CI} & \textbf{Tok/Q}\\
\midrule
SWC-Temporal   & 0.468 & [0.444,\,0.495] & 4\,257\\
SWC-Full       & 0.379 & [0.354,\,0.404] & 4\,130\\
Sliding window & 0.238 & [0.216,\,0.260] & 4\,125\\
Truncation     & 0.171 & [0.153,\,0.190] & 3\,662\\
\bottomrule
\end{tabular}
\end{table}

CIs are non-overlapping between SWC conditions and baselines. SWC-Temporal achieves
0.468, 0.298 above truncation and 0.231 above sliding window.

\subsection{Results by Category}
\label{sec:bycat}

Table~\ref{tab:bycategory} presents judge accuracy by category.

\begin{table*}[t]
\centering\small
\caption{Judge accuracy by category, matched question set. Category~2 (temporal)
highlighted.}
\label{tab:bycategory}
\renewcommand{\arraystretch}{1.1}
\begin{tabular}{lcccccccc}
\toprule
& \multicolumn{2}{c}{\textbf{SWC-Temporal}} & \multicolumn{2}{c}{\textbf{SWC-Full}}
& \multicolumn{2}{c}{\textbf{Sliding}} & \multicolumn{2}{c}{\textbf{Truncation}}\\
\cmidrule(lr){2-3}\cmidrule(lr){4-5}\cmidrule(lr){6-7}\cmidrule(lr){8-9}
\textbf{Category} & Acc & 95\% CI & Acc & 95\% CI & Acc & 95\% CI & Acc & 95\% CI\\
\midrule
1 Multi-hop   & 0.429 & [0.371,\,0.489] & 0.407 & [0.354,\,0.464] & 0.271 & [0.221,\,0.325] & 0.157 & [0.114,\,0.200]\\
\textbf{2 Temporal} & \textbf{0.470} & \textbf{[0.416,\,0.524]} & \textbf{0.156} & \textbf{[0.117,\,0.197]} & 0.171 & [0.130,\,0.213] & 0.137 & [0.102,\,0.178]\\
3 Open-domain & 0.427 & [0.333,\,0.521] & 0.354 & [0.260,\,0.448] & 0.354 & [0.260,\,0.458] & 0.323 & [0.229,\,0.427]\\
4 Single-hop  & 0.486 & [0.453,\,0.517] & 0.459 & [0.425,\,0.493] & 0.238 & [0.211,\,0.267] & 0.170 & [0.144,\,0.198]\\
\bottomrule
\end{tabular}
\end{table*}

\textbf{Paired delta, category~2:}
SWC-Temporal $-$ SWC-Full $= \mathbf{+0.314}$ $[0.254, 0.375]$, $n=315$.
The direction is positive in all ten conversations (per-conv deltas: $+$0.433,
$+$0.160, $+$0.259, $+$0.379, $+$0.385, $+$0.291, $+$0.177, $+$0.341, $+$0.364,
$+$0.226; unweighted mean $+$0.301).

On category~2, SWC-Temporal's CI $[0.416, 0.524]$ does not overlap with SWC-Full
$[0.117, 0.197]$ or either baseline. On categories~1 and~4, both SWC conditions have
non-overlapping CIs with both baselines. Category~3 CIs overlap for all conditions; no
reliable difference can be concluded.

\subsection{Preservation Analysis}
\label{sec:pres}

Table~\ref{tab:preservation} shows verbatim preservation rates across all ten
conversations.

\begin{table}[h]
\centering\small
\caption{Verbatim preservation rates by content type, all ten conversations.}
\label{tab:preservation}
\begin{tabular}{lrrcc}
\toprule
\textbf{Type} & \textbf{Found} & \textbf{SWC-F} & \textbf{SWC-T} & \textbf{$\times$}\\
\midrule
Temporal     & 952   & 3.05\% & 62.39\% & 20\\
Named entity & 7\,559 & 8.03\% & 8.22\%  & 1.02\\
Event        & 7\,072 & 5.03\% & 5.61\%  & 1.11\\
\bottomrule
\end{tabular}
\end{table}

Temporal protection causes an approximately 20-fold increase in temporal preservation
while barely affecting named entity ($\times$1.02) and event preservation ($\times$1.11).
The gist abstraction was always preserving events and participants; it was selectively
omitting \emph{when} they occurred.

Table~\ref{tab:perconv} shows per-conversation category~2 results.

\begin{table}[h]
\centering\small
\caption{Category~2 (temporal) judge accuracy by conversation.
$\dagger$Unweighted per-conv mean; matched paired estimate is $+0.314$ $[0.254,
0.375]$ (Table~\ref{tab:bycategory}).}
\label{tab:perconv}
\begin{tabular}{crrrcc}
\toprule
\textbf{C} & \textbf{T} & \textbf{F} & \boldmath$\Delta$ & \textbf{Sl} & \textbf{Tr}\\
\midrule
0 & 0.595 & 0.162 & $+$0.433 & 0.216 & 0.189\\
1 & 0.440 & 0.280 & $+$0.160 & 0.520 & 0.320\\
2 & 0.370 & 0.111 & $+$0.259 & 0.074 & 0.111\\
3 & 0.541 & 0.162 & $+$0.379 & 0.135 & 0.081\\
4 & 0.615 & 0.231 & $+$0.385 & 0.115 & 0.115\\
5 & 0.333 & 0.042 & $+$0.291 & 0.208 & 0.208\\
6 & 0.265 & 0.088 & $+$0.177 & 0.147 & 0.088\\
7 & 0.512 & 0.171 & $+$0.341 & 0.195 & 0.171\\
8 & 0.667 & 0.303 & $+$0.364 & 0.061 & 0.111\\
9 & 0.290 & 0.065 & $+$0.226 & 0.097 & 0.065\\
\midrule
\textbf{Mean} & \textbf{0.463} & \textbf{0.162} & $\mathbf{+0.301}$$^\dagger$ & 0.177 & 0.146\\
\bottomrule
\end{tabular}
\end{table}

\section{Discussion}

\subsection{The Temporal Anchor Mechanism}

Three independent sources mutually confirm the mechanism. The category accuracy pattern
shows SWC-Full fails specifically on temporal questions (0.156) while performing well
on single-hop (0.459) and multi-hop (0.407) --- consistent with selective, not general,
failure. The preservation analysis quantifies the selectivity: temporal preservation
jumps 20-fold (3.05\%$\to$62.39\%) while entities and events barely change. The
ablation closes the argument: $+0.314$ judge accuracy (paired, $n=315$) from one added
sentence, with non-overlapping CIs against SWC-Full, without materially reducing
multi-hop or single-hop performance.

\subsection{Temporal, Not Episodic}

SWC-Full scores 0.459 on single-hop factual questions --- specific facts from specific
conversational turns, episodic in the conventional sense. The framework does not fail
on episodic content generally; it fails on temporal information. The abstraction prompt
covers facts, relationships, decisions, and plans --- none of these are temporal anchors.
Temporal anchors fall outside all protected categories by default. SWC-Temporal corrects
this precisely.

\subsection{Open-Domain Results}

Category~3 CIs overlap completely across all conditions. Commonsense inference questions
depend substantially on model prior knowledge rather than compression strategy, making
them less sensitive to what is preserved. No conclusion can be drawn.

\subsection{Multi-hop and Single-hop Advantages}

Both SWC conditions have non-overlapping CIs with both baselines on categories~1 and~4.
SWC conditions score 0.407--0.429 on multi-hop against baselines of 0.157--0.271, and
0.459--0.486 on single-hop against 0.170--0.238. These gains are measured over all ten
conversations and reflect the advantage of preserving relational and biographical
structure over simple truncation.

\subsection{Conversation~1 Outlier}

Sliding window (0.520) exceeds SWC-Temporal (0.440) on category~2 in conversation~1.
The SWC-T vs SWC-F delta is still $+$0.160. Conversation~1 has only 25 temporal
questions and sliding window's 0.520 is anomalous relative to its 0.061--0.216 range
on the other nine conversations. Small cell sizes explain the deviation.

\subsection{Implications}

Any memory system using gist abstraction should explicitly protect temporal anchors.
The fix is a prompt modification, not an architectural change.

The category-level evaluation methodology and the preservation rate analysis together
provide a more informative picture than aggregate accuracy alone. Systems that achieve
high aggregate scores may mask systematic failures on specific question types.

\subsection{Limitations}

\textbf{Coverage:} Matched intersection covers all ten conversations (1\,935 text-only
questions; primary aggregate uses 1\,501 categories~1--4).
\textbf{Full context:} Evaluated on conversations~0--1 only; reference ceiling.
\textbf{Preservation metric:} Verbatim rate is a conservative proxy and may undercount
semantically faithful paraphrases.
\textbf{Transfer:} Temporal protection tested only in SWC; whether the same fix
improves sliding-window summarisation is left to future work.
\textbf{Open-domain:} Category~3 CIs overlap; no reliable conclusion.
\textbf{Scheduling:} Cannot be tested on an offline benchmark.

\section{Conclusion}

Generic gist abstraction preserves relational, event, and factual content while
selectively losing temporal anchors, because they are absent from standard prompt
specifications.
A preservation analysis confirms this: only 3.05\% of temporal expressions survive
verbatim in generic summaries, compared to $\sim$8\% for named entities and $\sim$5\%
for event mentions.

A one-sentence prompt modification recovers $+0.314$ $[0.254, 0.375]$ judge accuracy
on category-2 (temporal) questions (matched set, direction confirmed in all ten
conversations), without materially reducing multi-hop or single-hop performance.
The temporal protection is a targeted fix, not a general compression relaxation.

Temporal anchors should be treated as a protected category in any gist compression
component. Category-level evaluation paired with preservation rate analysis provides a
more complete picture of what a memory strategy preserves --- and what it leaves behind.

\section*{Acknowledgements}
The authors used Claude (Anthropic) for assistance with literature review, drafting, and
experimental design.

\bibliography{sleeping_agent}

@inproceedings{maharana2024evaluating,
  title={Evaluating Very Long-Term Conversational Memory of {LLM} Agents},
  author={Maharana, Adyasha and Lee, Dong-Ho and Tulyakov, Sergey and Bansal, Mohit and Barbieri, Francesco and Fang, Yuwei},
  booktitle={Proceedings of the 62nd Annual Meeting of the Association for Computational Linguistics},
  pages={13851--13870},
  year={2024}
}

@article{liu2024lost,
  title={Lost in the Middle: How Language Models Use Long Contexts},
  author={Liu, Nelson F. and Lin, Kevin and Hewitt, John and Paranjape, Ashwin and Bevilacqua, Michele and Petroni, Fabio and Liang, Percy},
  journal={Transactions of the Association for Computational Linguistics},
  volume={12},
  pages={157--173},
  year={2024}
}

@article{wang2025unable,
  title={Unable to Forget: Proactive Interference Reveals Working Memory Limits in {LLMs} Beyond Context Length},
  author={Wang, Chupei and Sun, Jiaqiu Vince},
  journal={arXiv preprint arXiv:2506.08184},
  year={2025}
}

@article{fang2025lightmem,
  title={{LightMem}: Lightweight and Efficient Memory-Augmented Generation},
  author={Fang, Jizhan and Deng, Xinyu and Xu, Huajun and Jiang, Zengwei and Tang, Yubo and Xu, Zhengyan and Deng, Shumin and Yao, Yuan and Wang, Mengdi and Qiao, Shuaiyi and Chen, Huajun and Zhang, Ningyu},
  journal={arXiv preprint arXiv:2510.18866},
  year={2025}
}

@article{chhikara2025mem0,
  title={Mem0: Building Production-Ready {AI} Agents with Scalable Long-Term Memory},
  author={Chhikara, Prateek and Khant, Dev and Aryan, Saket and Singh, Taranjeet and Yadav, Deshraj},
  journal={arXiv preprint arXiv:2504.19413},
  year={2025}
}

@article{wang2026memmachine,
  title={{MemMachine}: A Ground-Truth-Preserving Memory System for Personalized {AI} Agents},
  author={Wang, Sherry and Yu, Elaine and Love, Oliver and Zhang, Tina and Wong, Tony and Scargall, Sean and Fan, Cheng},
  journal={arXiv preprint arXiv:2604.04853},
  year={2026}
}

@article{patel2025engram,
  title={{ENGRAM}: Effective, Lightweight Memory Orchestration for Conversational Agents},
  author={Patel, Daivik and Patel, Shrenik},
  journal={arXiv preprint arXiv:2511.12960},
  year={2025}
}

@article{cruz2025afm,
  title={Adaptive Focus Memory for Language Models},
  author={Cruz, Christopher},
  journal={arXiv preprint arXiv:2511.12712},
  year={2025}
}

@article{kang2025acon,
  title={{ACON}: Optimizing Context Compression for Long-horizon {LLM} Agents},
  author={Kang, Minseok and others},
  journal={arXiv preprint arXiv:2510.00615},
  year={2025}
}

@article{xie2026sleepgate,
  title={Learning to Forget: Sleep-Inspired Memory Consolidation for Resolving Proactive Interference in Large Language Models},
  author={Xie, Ying},
  journal={arXiv preprint arXiv:2603.14517},
  year={2026}
}

@article{mcclelland1995complementary,
  title={Why There Are Complementary Learning Systems in the Hippocampus and Neocortex},
  author={McClelland, James L. and McNaughton, Bruce L. and O'Reilly, Randall C.},
  journal={Psychological Review},
  volume={102},
  number={3},
  pages={419--457},
  year={1995}
}

@article{tononi2006sleep,
  title={Sleep Function and Synaptic Homeostasis},
  author={Tononi, Giulio and Cirelli, Chiara},
  journal={Sleep Medicine Reviews},
  volume={10},
  number={1},
  pages={49--62},
  year={2006}
}

\appendix
\renewcommand{\thesection}{\Alph{section}}
\setcounter{section}{0}

\section{Worked Gist Abstraction Example}
\label{app:example}

\textbf{Original session:} \textit{``I finally went to the LGBTQ support group last
night --- 7~May. It was really moving. I spoke for about ten minutes about my experience
coming out to my parents.''}

\textbf{SWC-Full:} \textit{``Caroline attended an LGBTQ support group and shared her
coming-out experience, finding it emotionally meaningful.''}

\textbf{SWC-Temporal:} \textit{``On 7~May, Caroline attended an LGBTQ support group
and spoke for ten minutes about her coming-out experience, finding it emotionally
meaningful.''}

SWC-Full loses the date (7~May) and duration (ten minutes). Named entities and the
event are preserved in both; only temporal anchors differ.

\section{Baseline Results Across All Ten Conversations}
\label{app:baselines}

\begin{table}[h]
\centering\small
\caption{Overall judge accuracy, all ten conversations (includes category~5;
Table~\ref{tab:aggregate} uses categories~1--4 only).}
\label{tab:baseline_all}
\begin{tabular}{rrrr}
\toprule
\textbf{Conv} & \textbf{Q} & \textbf{Trunc} & \textbf{Sliding}\\
\midrule
0 & 197 & 0.355 & 0.386\\
1 & 102 & 0.422 & 0.461\\
2 & 191 & 0.298 & 0.330\\
3 & 237 & 0.304 & 0.354\\
4 & 240 & 0.317 & 0.329\\
5 & 158 & 0.367 & 0.367\\
6 & 189 & 0.249 & 0.344\\
7 & 233 & 0.283 & 0.343\\
8 & 192 & 0.297 & 0.328\\
9 & 196 & 0.260 & 0.316\\
\midrule
\textbf{Pooled} & \textbf{1\,935} & \textbf{0.305} & \textbf{0.347}\\
\bottomrule
\end{tabular}
\end{table}

Sliding window matches or exceeds truncation across all ten conversations
(conversation~5 is a tie at 0.367).

\section{SWC-Reactive Results}
\label{app:reactive}

SWC-Reactive (token-threshold triggering only, API default temperature,
conversations~0--2) achieved aggregate judge accuracy of 0.452 vs SWC-Full's 0.462.
Scheduling differences are not a validated contribution of this paper.

\section{Category~5 (Adversarial) Results}
\label{app:adversarial}

Category~5 questions are designed to be unanswerable; correct responses contain ``no
information available.'' Conditions removing more content score higher (truncation:
0.72--0.86 across conversations). This reflects content retention, not memory quality.

\section{Judge Prompt and Validation}
\label{app:judge}

\textbf{Judge prompt (temperature~0):}
\begin{quote}\small\itshape
Question: \{question\}. Gold answer: \{gold\_answer\}. Model answer: \{model\_answer\}.
Does the model answer correctly answer the question, given the gold answer?
Reply only YES or NO.
\end{quote}

\textbf{Validation:} 60-question blind sample; 95\% agreement (57/60); 100\% on
category~2 (17/17); all 3 disagreements conservative.

\textbf{Category examples:}
\textbf{Cat.~1} (multi-hop): ``What fields would Caroline likely pursue?''
\textbf{Cat.~2} (temporal): ``When did Caroline go to the LGBTQ support group?''
\textbf{Cat.~3} (open-domain): ``Would Caroline likely have Dr Seuss books?''
\textbf{Cat.~4} (single-hop): ``What did the charity race raise awareness for?''
\textbf{Cat.~5} (adversarial): Unanswerable questions; correct response acknowledges lack of information.

\end{document}